\documentclass[10pt,twocolumn,letterpaper]{article}

\usepackage[pagenumbers]{cvpr}      

\usepackage[T1]{fontenc}
\usepackage{multirow}       
\usepackage{makecell}       
\usepackage{tabularx}       
\usepackage{nicefrac}       
\usepackage{microtype}      
\newcolumntype{Y}{>{\centering\arraybackslash}X} 

\definecolor{revnew}{RGB}{0,150,0}
\definecolor{revmod}{RGB}{204,160,0}
\newif\ifrevmarks
\revmarkstrue
\ifrevmarks
  \newcommand{\addcolor}{\color{revnew}}
  \newcommand{\chgcolor}{\color{revmod}}
  \DeclareCaptionFont{addcap}{\color{revnew}}
  \DeclareCaptionFont{chgcap}{\color{revmod}}

\else
  \newcommand{\addcolor}{}
  \newcommand{\chgcolor}{}

\fi

\definecolor{cvprblue}{rgb}{0.21,0.49,0.74}
\usepackage[pagebackref,breaklinks,colorlinks,allcolors=cvprblue]{hyperref}

\def\confName{CVPR}
\def\confYear{2027}

\title{EATS: Scaling Tool-Use Experience for Medical Reasoning Agents}

\author{%
Mingyuan Li$^{1}$, Tong Jia$^{1}$, Haoran Wang$^{1}$, Wenbo Qi$^{2}$,\\
Ziyuan Yang$^{2}$, Bowen Ma$^{1}$, Jie Gui$^{3}$, Baosheng Yu$^{2}$\\[4pt]
{\small $^{1}$College of Information Science and Engineering, Northeastern University, China}\\
{\small $^{2}$Nanyang Technological University\qquad $^{3}$Southeast University}\\[4pt]
{\small Tong Jia: \href{mailto:jiatong@ise.neu.edu.cn}{\texttt{jiatong@ise.neu.edu.cn}}\qquad
Baosheng Yu: \href{mailto:baosheng.yu@ntu.edu.sg}{\texttt{baosheng.yu@ntu.edu.sg}}}
}

\hypersetup{hypertexnames=false,
  pdftitle={EATS: Scaling Tool-Use Experience for Medical Reasoning Agents},
  pdfauthor={Mingyuan Li, Tong Jia, Haoran Wang, Wenbo Qi, Ziyuan Yang, Bowen Ma, Jie Gui, Baosheng Yu}}
\begin{document}

\maketitle

\begin{abstract}
Medical reasoning agents combine large language model (LLM) reasoning with clinical workflows and specialized neural tools. However, general-purpose LLMs receive limited training exposure to domain-specific medical tools, and the uncertainty and distributional shifts of neural tool outputs make selecting tools from simple descriptions significantly more challenging than in deterministic settings such as calculators or dictionary lookups. In this paper, we propose EATS, an experience-augmented tool selection strategy that leverages accumulated tool-use trajectories to improve tool selection in medical reasoning agents. EATS constructs a memory of past interactions, including queries, tool choices, arguments, and self-evaluated feedback, and retrieves relevant experiences to guide tool selection for new queries. As memory construction does not require ground-truth supervision, EATS scales naturally with test-time experience. We evaluate EATS on tool-augmented medical reasoning tasks, including closed-ended visual question answering and free-form report generation. Experiments on four public datasets, ChestAgentBench, CheXbench, MIMIC-CXR, and SLAKE-VQA, across three multimodal LLMs, Qwen3-VL 2B, Qwen3-VL 32B, and GPT-5.5, show consistent gains in tool selection behavior and downstream medical reasoning performance. Code is available at \url{https://github.com/Limingyuan001/EATS}.
\end{abstract}

\section{Introduction}
\label{sec:introduction}

Medical reasoning agents integrate large language model (LLM)–based reasoning with clinical workflows and specialized neural tools, allowing perception, analysis, and decision-making to be jointly carried out across complex diagnostic tasks~\citep{zhou2026medicalsurvey}. In practice, these agents operate through an iterative plan–act loop~\cite{yao2022react}, where each reasoning step may trigger the use of external tools to supplement the limited parametric knowledge of foundation models~\cite{fallahpour2025medrax,li2024mmedagent}. For example, in medical image diagnosis, an agent may invoke neural networks for tasks such as classification, segmentation, and report generation, thereby extending its perceptual and analytical capacity. However, integrating medical tools introduces additional challenges. Unlike deterministic tools with stable, predictable outputs, medical models often are not universally optimal across all cases. This variability reduces the reliability of tool selection and complicates coordination across multiple steps. As the reasoning process unfolds, errors introduced at earlier stages can propagate through subsequent steps, leading to compounding inaccuracies in downstream decisions. Consequently, selecting appropriate tools at each stage remains a central challenge in developing effective medical reasoning agents.

\begin{figure*}[t]
    \centering
    \captionsetup{skip=7.5pt}
    \includegraphics[width=\textwidth]{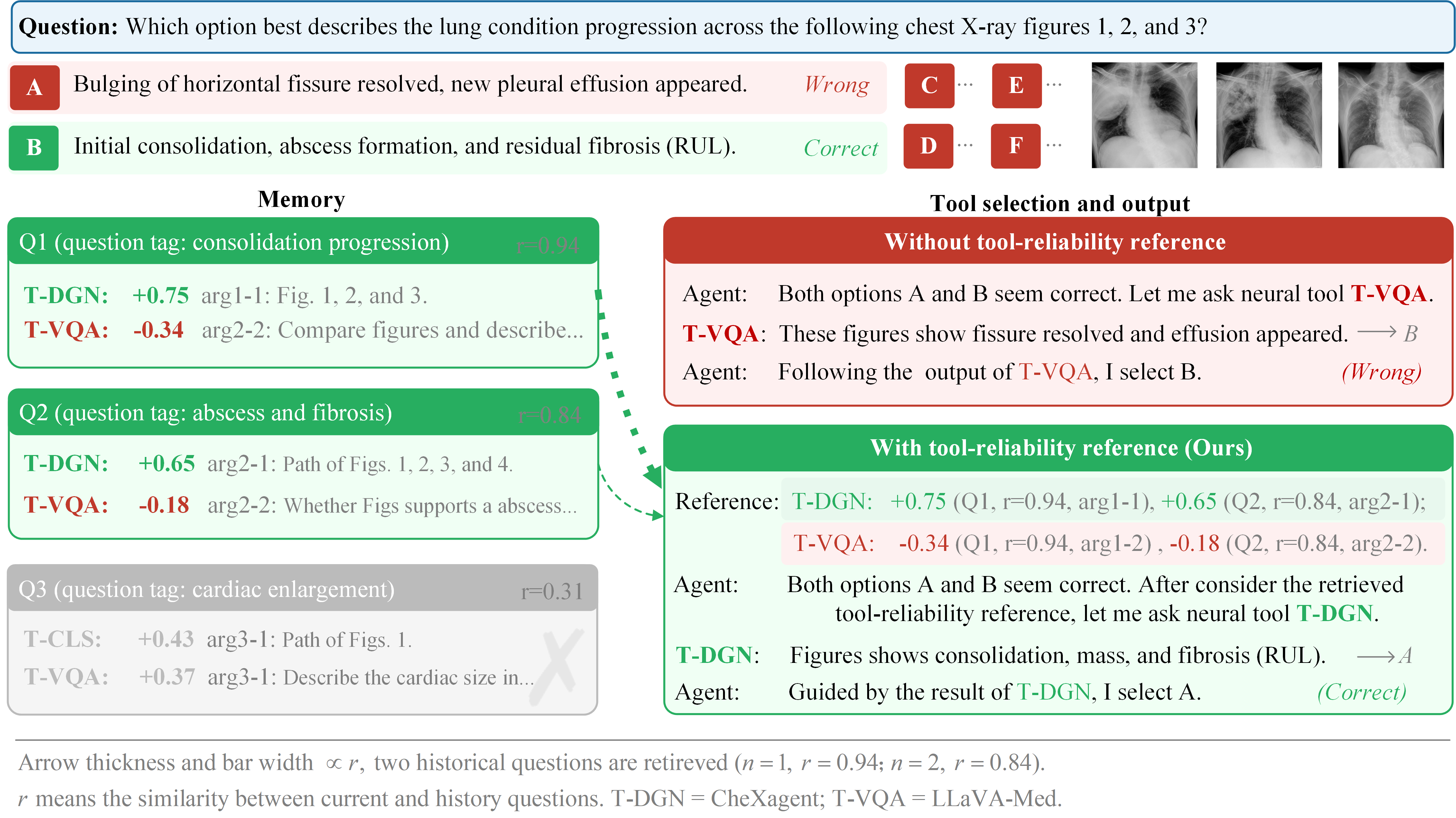}
    \caption{\textbf{Illustration of tool-use experience for tool selection.} Without tool-use references (e.g., tool reliability and case similarity), the ReAct-style agent~\cite{yao2022react,fallahpour2025medrax} selects T-VQA~\cite{li2023llavamed} and produces an incorrect answer. With tool-reliability information retrieved from similar past cases, it selects T-DGN~\cite{chen2024chexagent} and produces the correct answer.}
\vspace{2pt}
\label{fig:problem}
\end{figure*}

Current tool-selection strategies in medical reasoning agents typically match user queries with tool descriptions that emphasize predefined functions or capabilities~\cite{li2024mmedagent,fallahpour2025medrax}, limiting their ability to incorporate patient-specific context. While this paradigm is effective for deterministic tools such as calculators or search engines in general-agent systems, it is less suitable for medical settings where neural tools exhibit output uncertainty and case-dependent variability~\cite{li2023llavamed, chen2024chexagent}. Recent approaches explore retraining-based methods to improve tool selection~\cite{schick2023toolformer, yang2026tooltree, liu2026toolrla, shinn2023reflexion}, but their scalability in clinical domains remains constrained by data heterogeneity, privacy restrictions, and the black-box nature of commercial LLMs.  In contrast, similar or relevant cases can serve as empirical references for tool selection. They connect the current patient context with prior tool-use outcomes, allowing tools that performed well in comparable cases to be preferentially selected, while less reliable ones can be avoided. This case-based grounding also helps identify recurring failure patterns across steps, thereby reducing error accumulation in multi-stage reasoning and supporting more adaptive and context-aware tool selection. As shown in Figure~\ref{fig:problem}, without tool-use experience, the agent selects a plausible but inappropriate tool and yields an incorrect diagnosis; with tool-use experiences from two similar cases, the agent identifies a more reliable tool and produces the correct diagnosis.

In this paper, we propose EATS, an experience-augmented tool selection strategy that leverages accumulated tool-use trajectories to improve medical reasoning agents. EATS constructs a memory of past interactions, including queries, tool selections, input arguments, tool outputs, and self-evaluated feedback. The agent then retrieves relevant experiences and uses them as a tool-use reference to guide tool selection for new queries. These retrieved experiences are injected into the reasoning context, enabling reliable tool selection without additional training or retraining the model. After each case, an LLM-as-a-Judge module evaluates the completed tool-use trajectory and assigns scores to individual tool-use events based on multiple criteria across two dimensions, including clinical relevance, informational gain, diagnostic discriminability, citation fidelity, and answer consistency. The updated experiences are then stored in memory for future retrieval. Since memory construction does not rely on ground-truth supervision, EATS can continuously scale with test-time interactions and accumulated experience.

We evaluate the proposed method on four chest X-ray benchmarks: ChestAgentBench~\citep{fallahpour2025medrax}, CheXbench~\citep{chen2024chexagent}, MIMIC-CXR~\citep{johnson2019mimic}, and SLAKE-VQA~\citep{liu2021slake}, using three multimodal foundation models (Qwen3-VL 2B, Qwen3-VL 32B, and GPT-5.5). The method consistently outperforms the MedRAX~\citep{fallahpour2025medrax} baseline (without memory) across both small and large Qwen3-VL models and the frontier model GPT-5.5, with gains increasing as tool-use experience accumulates. The approach also generalizes to three additional reasoning benchmarks covering report generation and closed-ended VQA. Further analysis shows that performance improvements mainly arise from better tool-selection decisions, including which tool to use, when to use it, and how to set its arguments, rather than increased diversity of inputs or outputs.


\section{Related Work}
\label{sec:related}

\paragraph{Medical Reasoning Agents.}
Medical AI has evolved from fine-tuning static language and vision-language models~\citep{singhal2023large,li2023llavamed,chen2024chexagent} to agent-based systems that dynamically invoke external tools for multi-step clinical reasoning~\citep{zhou2026medicalsurvey}. Early models achieved strong performance on clinical question answering but remained passive, lacking the ability to actively coordinate tool usage during reasoning~\citep{li2023llavamed,chen2024chexagent,chambon2024chexpert}. Recent approaches introduce tool-augmented agents with explicit planning and tool calling~\citep{yao2022react,fallahpour2025medrax,lou2026cxragent}, including multi-role LLM ensembles~\citep{tang2024medagents,kim2024mdagents}, instruction-tuned unified models for multimodal tool use~\citep{li2024mmedagent}, expanded tool libraries for imaging tasks~\citep{fallahpour2025medrax}, and retrieval-based workflow planning or rare-disease reasoning~\citep{wang2026medagentpro,zhao2026agentic,chen2026rareagents}. Despite these advances, existing methods do not explicitly support self-improvement from accumulated tool-use experience and typically require retraining or external supervision to improve over time.

\paragraph{Tool Use \& Memory.}
General-agent systems have explored multiple approaches for tool use. Toolformer~\citep{schick2023toolformer} enables self-supervised learning of tool-call decisions, while ReAct~\citep{yao2022react} integrates reasoning traces and tool actions within a unified interaction loop. More structured methods include hierarchical tool organization~\citep{yang2026tooltree}, utility-based pruning of tool invocations~\citep{liu2026utility}, and reinforcement learning for tool-selection policies~\citep{liu2026toolrla}. Benchmarks for tool-use evaluation have also been proposed~\citep{xu2026evolution,zhao2026ama,lu2026medvistagym}. However, these methods are primarily designed for tools with well-defined and static capabilities, whereas medical neural tools exhibit case-dependent reliability that cannot be inferred from descriptions alone, and existing approaches do not address this issue without retraining. In parallel, external memory has been widely used to enhance agent reasoning~\citep{du2026memorysurvey}. Existing memory mechanisms include long-term memory, which stores persistent knowledge across tasks~\citep{zhong2024memorybank,jiang2024long}, and short-term memory, which records in-context information such as inputs, intermediate reasoning, tool calls, and outputs~\citep{yao2022react,wu2025human}. Several methods store and refine agent experiences, including verbal self-reflection~\citep{shinn2023reflexion}, episodic tool-use memories~\citep{xiao2025toolmem,wang2026veriagent,yuan2026ta}, and unified memory management across short- and long-term stores~\citep{yu2026agentic,chang2026memcollab}. Some studies further use memory to support reinforcement learning-based fine-tuning~\citep{schick2023toolformer,wang2025reinforcement,acikgoz2025self,yan2025memory}, improving tool-selection policies but requiring retraining and not leveraging memory directly at inference time.

\section{Method}
\label{sec:method}

This section first formulates the medical reasoning agent and provides an overview of the proposed EATS strategy. It then details the key components, including memory construction, tool selection, and memory updating.

\subsection{Problem Formulation}
\label{sec:problem}

We consider a ReAct-style agent for chest X-ray interpretation equipped with $K$ specialized neural tools $\mathcal{T}=\{T_1,\dots, T_K\}$. Each input is a clinical case $\mathcal{X}=\{I, Q\}$, consisting of chest X-ray image(s) $I$ and a clinical question $Q$. A multimodal foundation model $f_{\theta}$ serves as the agent’s reasoning core, conditioned on a system prompt $P_{\mathrm{syst}}$ that defines tool descriptions and usage instructions.
At each step $l \in \{1,\dots,L\}$, the agent generates a reasoning message $C_l$ and selects a subset of tools $\mathcal{T}_l \subseteq \mathcal{T}$ to produce feedback $E_l$:
\begin{align}
\label{eq:react}
C_l,\;\mathcal{T}_l
&=
f_{\theta}\!\left(\mathcal{X},\,C_{<l},\,E_{<l};\,P_\mathrm{syst}\right),\\
E_l &= \{\,T(\mathcal{X},C_l)\mid T\in\mathcal{T}_l\,\},
\end{align}
where $C_{<l}$ and $E_{<l}$ denote prior reasoning and tool outputs. The process terminates when $\mathcal{T}_l=\emptyset$, and the final prediction is $\hat{y}=C_L$. Unlike deterministic tools such as calculators or search engines, medical neural tools (e.g., LLaVA-Med, CheXagent) exhibit case-dependent behavior and vary in reliability across inputs. Therefore, the mapping from $\{\mathcal{X}, C_{<l}, E_{<l}\}$ to appropriate tool selection $\mathcal{T}_l$ cannot be reliably captured by static tool descriptions in $P_{\mathrm{syst}}$ alone.

\begin{figure*}[t]
    \centering
    \includegraphics[width=\textwidth]{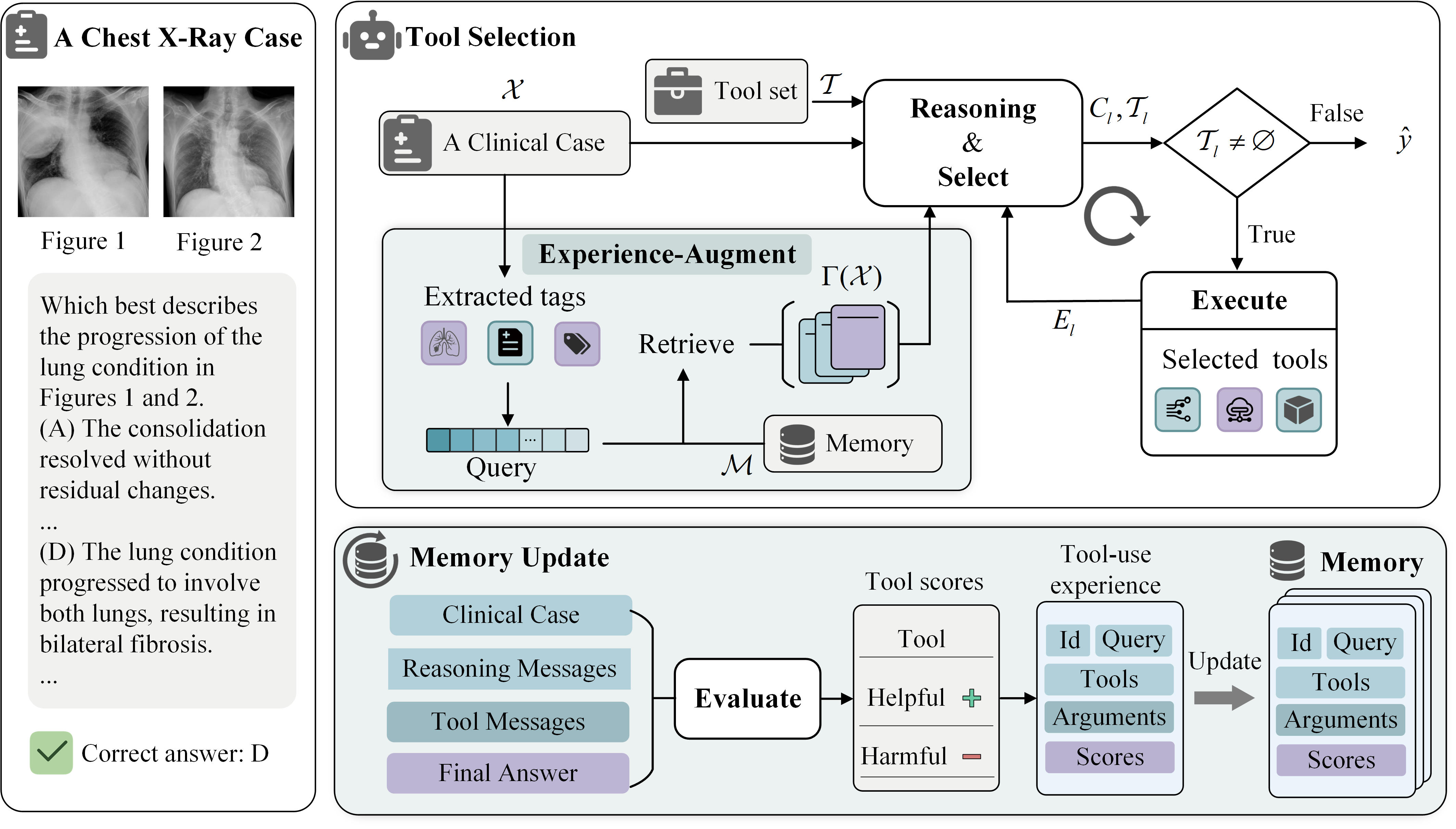}
    \caption{Overview of the proposed EATS.
\textbf{(Top)} Tool Selection: the agent retrieves similar historical entries from memory $\mathcal{M}$ to assemble tool-reliability reference $\Gamma(\mathcal{X})$ as additional context, and iteratively performs reasoning and tool selection until $\mathcal{T}_l=\emptyset$.
\textbf{(Bottom)} Memory Update: the completed trajectory is evaluated across five criteria to score each tool event, and the resulting
tool-use experience $\mathcal{U}$ is written back into $\mathcal{M}$ for future retrieval.
\textbf{(Left)} A representative CXR case.
}
\label{fig:agent}
\end{figure*}

As illustrated in Figure~\ref{fig:agent}, we propose an experience-augmented tool selection strategy for medical reasoning agents as follows. Given a new case $\mathcal{X}$, the agent retrieves similar historical cases from memory $\mathcal{M}$ and incorporates their tool-reliability references $\Gamma(\mathcal{X})$ into the foundation model context to support reliability-aware decision-making. After inference, the contribution of each tool invocation is evaluated for the current case, and the resulting experience is stored in memory as a new tool-use record $\mathcal{U}$ together with the case representation $\mathbf{z}$ for future retrieval. The following sections detail each component.

\subsection{Experience Memory}
\label{sec:memory}

The first component of our strategy is a structured experience memory that stores historical cases and their associated tool-use experiences in a retrievable form. Instead of retaining only raw queries or final predictions, the memory preserves information most relevant for future tool selection, including retrieval query embeddings, invoked tools, and their effectiveness scores. Formally, the memory is organized as a key–value store as follows:
\begin{equation}
\mathcal{M} = \bigl\{\,(\mathbf{z}_n,\;\mathcal{U}_n)\,\bigr\}_{n=1}^{|\mathcal{M}|},
\end{equation}
where $\mathbf{z}_n\in\mathbb{R}^{384}$ is the embedding key for the $n$-th case $\mathcal{X}_n=\{I_n, Q_n\}$, computed in the phrase of tool selection. The corresponding value $\mathcal{U}_n=\{u_{n,1}, \dots, u_{n, J_n}\}$ stores a set of $J_n$ tool events collected from the completed reasoning trajectory, Each event is defined as
\begin{equation}
u_{n,j} =
\bigl(
T_{n,j},\;
a_{n,j},\;
s_{n,j}
\bigr).
\end{equation}
where $T_{n,j}\in\mathcal{T}$ denotes the invoked tool, $a_{n,j}$ its input arguments (e.g., prompts for LLaVA-Med-style tools), and $s_{n,j}\in[-1,1]$ a scalar score measuring the utility of that tool invocation for solving case $n$. This score serves as the supervision signal stored in memory for future retrieval and decision-making. The estimation of $s_{n,j}$ is described in Section~\ref{sec:eval}.

\subsection{Tool Selection}
\label{sec:inference}

At inference time, the agent processes a new case $\mathcal{X}=\{I,Q\}$ by leveraging historical tool-use experiences retrieved from the experience memory. We first extract structured semantic tags from $Q$ and encode them into a query vector:
\begin{equation}
\mathbf{z} = \phi(\mathbf{c}) \in \mathbb{R}^{384},
\quad
\mathbf{c} = f_{\theta}(Q;\,P_{\mathrm{ext}}),
\end{equation}
where $P_{\mathrm{ext}}$ is a prompt for tag extraction (see Appendix~\ref{sec:prompt_extract}), $\mathbf{c}$ contains six types of structured tags, and $\phi(\cdot)$ is a text embedding model~\cite{xiao2024c}.
For each memory entry $(\mathbf{z}_n, \mathcal{U}_n)\in\mathcal{M}$, we compute a similarity score
\begin{equation}
r_n = \cos(\mathbf{z},\mathbf{z}_n), \quad n \in \{1,\dots,|\mathcal{M}|\},
\end{equation}
which measures the semantic similarity between the current case and historical cases. We then construct a retrieved set by selecting entries with similarity above a threshold $\tau$ and retaining at most $k$ highest-scoring items:
\begin{align}
\mathcal{R}_{\tau,k}(\mathcal{X};\,\mathcal{M})
&\subseteq \bigl\{\, n \mid r_n \ge \tau \,\bigr\},\\
|\mathcal{R}_{\tau,k}(\mathcal{X};\,\mathcal{M})| &\le k,
\end{align}
where entries are ranked by $r_n$ in descending order and only the top $k$ are retained, $\tau$ is a similarity threshold, and $k$ is the maximum number of retrieved entries.
If no historical entry satisfies the threshold, $\mathcal{R}_{\tau,k}(\mathcal{X};\,\mathcal{M})=\emptyset$ and the agent reduces to plain prompt-driven reasoning without memory support.
For each retrieved index $n\in\mathcal{R}_{\tau,k}(\mathcal{X};\,\mathcal{M})$, the corresponding tool-use experience $\mathcal{U}_n$ and similarity score $r_n$ are assembled into a tool-reliability reference:
\begin{equation}
\Gamma(\mathcal{X})
= \bigl\{\,(r_n,\,\mathcal{U}_n)\mid n \in \mathcal{R}_{\tau,k}(\mathcal{X};\,\mathcal{M})\,\bigr\},
\end{equation}
which summarizes tool behaviors and their effectiveness in similar cases.
Conditioned on the current case, reasoning history, and $\Gamma(\mathcal{X})$, the agent performs iterative experience-augmented reasoning. At each step $l\in\{1,\dots,L\}$:
\begin{align}
C_l,\;\mathcal{T}_l
&=
f_{\theta}\!\left(
\mathcal{X},C_{<l},E_{<l},\Gamma(\mathcal{X}); P_{\mathrm{syst}}
\right),\\
E_l &= \{\,T(\mathcal{X},C_l)\mid T\in\mathcal{T}_l\,\}.
\end{align}
The process terminates when $\mathcal{T}_l = \emptyset$ or a maximum of $L=7$ steps is reached, and the final reasoning output is taken as the prediction $\hat{y}$.

\subsection{Memory Update}
\label{sec:eval}

\paragraph{Evaluate.}
As shown in Figure~\ref{fig:agent} (bottom), we evaluate each tool event in the completed trajectory using a process-based protocol that does not require ground-truth labels (a comparison with a ground-truth-based judge is provided in Appendix~\ref{sec:judge_validation}). Guided by $P_{\mathrm{eval}}$, the same foundation model $f_{\theta}$ assigns scores to each event using LLM-as-a-Judge with five criteria that assess both output quality and its role in the reasoning process. The criteria are:
(1) \textit{clinical relevance}, whether the tool output addresses the relevant pathology or task;
(2) \textit{informational increment}, whether it provides information beyond the input question;
(3) \textit{diagnostic discriminability}, whether it helps distinguish among competing diagnoses;
(4) \textit{citation fidelity}, whether the reasoning properly uses or references the tool output; and
(5) \textit{answer consistency}, whether the output is consistent with the final prediction.
The first three criteria mainly evaluate the intrinsic quality of the tool output, while the last two assess how coherently it is integrated into the reasoning process. Each criterion is scored independently on a five-point scale mapped to $[-1, 1]$; the full rubric and prompt template $P_{\mathrm{eval}}$ are provided in Appendix~\ref{sec:prompt_eval}. For each tool event $j \in \{1,\dots,J\}$, the contribution score is defined as:
\begin{equation}
  \label{equation_score}
s_j
=
\mathrm{M}\!\left(
f_{\theta}\!\left(
\mathcal{X},C_{<L},E_{<L},\hat{y};\,P_{\mathrm{eval}}
\right)
\right)\in[-1,1],
\end{equation}
where $\mathrm{M}(\cdot)$ denotes the operator that computes the arithmetic mean of the five criterion scores.

\paragraph{Update.}
The experience memory is updated with the newly collected tool-use experience and the corresponding query embedding for future retrieval.
Specifically, after all $J$ tool events in a case are scored, we construct the tool-use experience as a set of tuples $\mathcal{U}=\{(T_j,a_j,s_j)\}_{j=1}^{J}$. The query embedding $\mathbf{z}$, computed during inference, is reused as the key for memory storage. The resulting key–value pair $(\mathbf{z}, \mathcal{U})$ is then appended to the memory:
\begin{equation}
\mathcal{M} \leftarrow \mathcal{M} \cup \{(\mathbf{z}, \mathcal{U})\}.
\end{equation}
In this way, each processed case enriches the memory with evaluated tool-use experience, supporting more reliable tool selection and evidence integration in future cases.

\section{Experiments}
\label{sec:experiments}

In this section, we study three questions: (i) whether the proposed memory improves accuracy across foundation model scales and chest X-ray tasks; (ii) whether performance scales with accumulated tool-use experience; and (iii) how changes in tool-use behavior translate into accuracy gains. To isolate the effect of memory, our strategy shares the same inference loop and tool library as MedRAX~\citep{fallahpour2025medrax}; the only differences are the memory-augmented context $\Gamma(\mathcal{X})$ injected into the reasoning process and the post-hoc tool-contribution scoring used for memory updates. We therefore adopt the MedRAX agent, which is without memory, as the baseline for comparisons and evaluate on chest X-ray interpretation with established neural tools.


\subsection{Experimental Setup}
\label{sec:setup}

\textbf{Datasets.}
We evaluate on four chest X-ray benchmarks adopted by MedRAX~\citep{fallahpour2025medrax}, covering complementary aspects of clinical reasoning:
\begin{itemize}[leftmargin=*, nosep]
  \item \textbf{ChestAgentBench}~\citep{fallahpour2025medrax}: a comprehensive CXR reasoning benchmark of $2,500$ six-choice questions across seven competencies---detection, classification, localization, comparison, relationship, characterization, and diagnosis. Performance is measured by overall and per-category accuracy.
  \item \textbf{CheXbench}~\citep{chen2024chexagent}: we use the visual question answering subset (238 questions from Rad-Restruct~\citep{pellegrini2023rad} and SLAKE~\citep{liu2021slake}) and the fine-grained image-text reasoning subset (380 questions from OpenI~\citep{demner2016preparing}), as they most closely mirror complex clinical workflows that require precise differentiation between similar findings. We report accuracy on each subset and the overall mean.
  \item \textbf{MIMIC-CXR}~\citep{johnson2019mimic}: radiology report generation on $3,858$ single-image chest X-rays from the MIMIC-CXR test set. Following M4CXR~\citep{park2025m4cxr}, we report micro- and macro-averaged F1 over $14$ medical observation labels (mF1-14, MF1-14; CheXpert labels~\citep{irvin2019chexpert} extracted by CheXbert~\citep{smit2020combining}) and over $5$ key findings (mF1-5, MF1-5; cardiomegaly, edema, consolidation, atelectasis, and pleural effusion extracted by CheXbert~\citep{smit2020combining}).
  \item \textbf{SLAKE-VQA}~\citep{liu2021slake}: medical visual question answering on $114$ closed-ended English chest X-ray samples filtered from the SLAKE test set, following M4CXR~\citep{park2025m4cxr}. We report exact-match accuracy and recall (proportion of ground-truth words present in the generated response).
\end{itemize}

\begin{table*}[t]
\centering
\small
\setlength{\tabcolsep}{4pt}
\caption{The six specialized tools available to the agent. Implementations are identical to MedRAX~\citep{fallahpour2025medrax}; we add the abbreviations used throughout the analysis, together with the input arguments and output content of each tool.}
\label{tab:tools}
\begin{tabular*}{\linewidth}{@{\extracolsep{\fill}}lllll@{}}
\toprule
Abbr. & Task & Implementation & Input (\texttt{args}) & Output \\
\midrule
\texttt{Cls} & Classification    & XRV~\citep{cohen2022torchxrayvision}  & Image (\texttt{image\_path})                              & 18 pathology scores \\
\texttt{Dgn} & Diagnosis         & CheXagent~\cite{chen2024chexagent}     & Image (\texttt{image\_path}) + prompt                 & Findings / diagnosis \\
\texttt{Rpt} & Report Generation & CheXpert Plus~\citep{chambon2024chexpert} & Image (\texttt{image\_path})                              & Findings + impression \\
\texttt{Seg} & Segmentation      & PSPNet~\citep{lian2021structure}        & Image (\texttt{image\_path}) + organ list                 & Organ mask \\
\texttt{Vqa} & VQA             & LLaVA-Med~\cite{li2023llavamed}     & Question + optional image                                  & Free-text answer \\
\texttt{Grd} & Phrase Grounding  & Maira-2~\citep{bannur2024maira}       & Image (\texttt{image\_path}) + phrase                     & BBox coordinates \\
\bottomrule
\end{tabular*}
\end{table*}

\textbf{Settings.}
We designed two strategies for constructing the tool-use experience memory to assess the self-improving capability of our strategy on ChestAgentBench: a \textbf{test-time} setting and a \textbf{pre-construction} setting. In the test-time setting, the dataset was processed sequentially with dynamic memory: each case was first inferred against the existing memory, then its tool-use experience was scored and written back to the memory. We further parameterized this setting by a memory-coverage ratio $\rho{\in}[0,1]$: the first $\rho{\cdot}N$ cases of the dataset were processed with dynamic memory updates (read-then-write), while the remaining $(1{-}\rho){\cdot}N$ cases were processed against the resulting fixed memory (read-only); thus $\rho{=}0$ recovers the baseline agent MedRAX and $\rho{=}1$ corresponds to the full test-time setting where every case contributes to memory. In the pre-construction setting, inference uses a fixed, pre-constructed memory; for a controlled comparison, we adopted 5-fold cross-validation, in which four folds were used to populate the memory and the remaining fold was held out for evaluation. 

\textbf{Implementation Details.}
We evaluate the method using two open-source multimodal foundation models of different scales, Qwen3-VL 2B and Qwen3-VL 32B~\citep{bai2025qwen3}, and include GPT-5.5 as a proprietary frontier model to assess the performance of our memory-augmented strategy. Unless otherwise specified, Qwen3-VL 32B is used as the default core model. The agent is equipped with six specialized neural tools, one per task category (Table~\ref{tab:tools}). The tool library is identical to MedRAX~\citep{fallahpour2025medrax}, so performance differences are attributable to the proposed memory mechanism. All experiments are conducted on two NVIDIA RTX Pro 6000 GPUs (96GB VRAM each). We use \texttt{bge-small-en-v1.5}\citep{xiao2024c} ($d{=}384$) as the embedding model $\phi(\cdot)$ and cap the reasoning steps at $L{=}7$. For memory retrieval, we use top-$k{=}3$ with a similarity threshold $\tau{=}0.99$ by default; sensitivity is analyzed in Section~\ref{sec:hp_ablation}. The Qwen3-VL 2B and 32B models are deployed via local OpenAI-compatible endpoints with \texttt{qwen3-vl-2b-instruct-fp8} and \texttt{qwen3-vl-32b-instruct}, respectively, both with temperature $0.2$ and top-$p{=}0.95$. GPT-5.5 uses the default OpenAI endpoint with \texttt{gpt-5.5}, \texttt{service\_tier=flex}, temperature $0.2$, no top-$p$, and a $600$ seconds timeout. Full implementation details about system prompt and tool descriptions, tag extract prompt, and tool evaluation prompt are provided in Appendices~\ref{sec:implementation_details_app}, \ref{sec:prompt_extract}, and~\ref{sec:prompt_eval}.

\subsection{Main Results}
\label{sec:main_results}

We present main results in two complementary views: a cross-model comparison on ChestAgentBench (Table~\ref{tab:main_backbone}), and a memory-scaling curve on the same benchmark (Table~\ref{tab:memory_scale}). Together they support our central claim: \emph{it consistently improves agent accuracy across substantially different foundation model scales, and the gain grows monotonically as accumulated tool-use experience scales.}

\begin{table*}[t]
\centering
\small
\caption{
Model performance on ChestAgentBench~\cite{fallahpour2025medrax} across seven task categories. The \emph{upper block} lists standalone multimodal FMs and the prior agent reported in \cite{fallahpour2025medrax}. The \emph{middle block} lists the MedRAX agent (Baseline) re-run with Qwen3-VL 2B, Qwen3-VL 32B, and GPT-5.5, and the \emph{lower block} lists our memory-augmented agent with the same three core FMs. Each pair of rows sharing a core FM isolates the effect of memory.
}
\label{tab:main_backbone}
\setlength{\tabcolsep}{3pt}
\begin{tabular*}{\linewidth}{@{\extracolsep{\fill}}llcccccccc@{}}
\toprule
Core FM & Agent & Overall & Detect. & Classif. & Local. & Compar. & Relat. & Diagn. & Charact. \\
\midrule
\multicolumn{10}{l}{\textit{Reference: standalone multimodal FMs and prior agent}} \\
LLaVA-Med~\citep{li2023llavamed}    & ---       & 28.7 & 32.4 & 30.8 & 30.2 & 30.6 & 31.8 & 29.3 & 28.8 \\
CheXagent~\citep{chen2024chexagent} & ---       & 39.5 & 38.7 & 34.7 & 42.5 & 38.5 & 39.8 & 33.5 & 34.2 \\
Llama-3.2-90B                       & ---       & 57.9 & 58.1 & 56.5 & 59.9 & 57.5 & 59.3 & 55.9 & 58.0 \\
Qwen3-VL 32B~\citep{bai2025qwen3} &---& 63.1& 63.7 &62.5& 64.1& 63.8 &65.7& 61.1& 61.3\\
GPT-4o~\citep{hurst2024gpt}                              & ---       & 56.4 & 58.7 & 54.6 & 59.0 & 55.5 & 59.0 & 52.6 & 56.1 \\
GPT-5.5 & --- & 82.4 & 83.6 & 83.6 & 80.4 & 81.1 & 81.4 & 81.6 & 82.8 \\
GPT-4o~\citep{hurst2024gpt}                              & MedRAX~\citep{fallahpour2025medrax} & 63.1 & 64.1 & 62.9 & 63.6 & 61.8 & 63.1 & 62.5 & 64.0 \\
\midrule
\multicolumn{10}{l}{\textit{Baseline (w/o Memory)}} \\
Qwen3-VL 2B~\citep{bai2025qwen3}  & MedRAX~\citep{fallahpour2025medrax} & 56.8& 57.6 & 58.1 & 55.1 & 56.6 & 57.1 & 57.3 & 55.8    \\
Qwen3-VL 32B~\citep{bai2025qwen3} & MedRAX~\citep{fallahpour2025medrax} & 64.0 & 65.7 & 63.4 & 65.1 & 61.9 & 63.6 & 63.0 & 66.3 \\
GPT-5.5 & MedRAX~\citep{fallahpour2025medrax} & 83.1 & 84.0 & 84.2 & 81.6 & 81.8 & 82.1 & 81.9 & 83.7 \\
\midrule
\multicolumn{10}{l}{\textit{Ours (w/ Memory)}} \\
Qwen3-VL 2B~\citep{bai2025qwen3}  & EATS (Ours) & 57.6& 59.0 &58.9 &55.7 &56.0 &56.9 &58.3 &58.1           \\
Qwen3-VL 32B~\citep{bai2025qwen3} & EATS (Ours) & 65.2 & 67.4 & 65.3 & 65.2 & 63.4 & 65.2 & 64.3 & 66.1 \\
GPT-5.5      & EATS (Ours) & 85.7 & 86.7 & 86.8 & 84.0 & 84.5 & 84.7 & 86.6 & 86.2   \\
\bottomrule
\end{tabular*}
\end{table*}

\textbf{Foundation Model.}
Table~\ref{tab:main_backbone} reports overall and per-category accuracy on ChestAgentBench. EATS improves the matched MedRAX baseline with all three core FMs, by $+0.8$, $+1.2$, and $+2.6$ points for Qwen3-VL 2B, Qwen3-VL 32B, and GPT-5.5, respectively, demonstrating that the memory benefit persists across substantially different FM scales rather than being tied to a particular model size. With GPT-5.5, EATS reaches an overall accuracy of $85.7\%$, compared with $83.1\%$ for the memory-free agent and $82.4\%$ for the standalone FM. These results suggest that memory-guided tool selection remains useful across model scales, while stronger FMs further improve the performance of the proposed strategy.
The reference rows in the upper block of the table further confirm that the agent strategy itself is necessary to surpass single-model performance on ChestAgentBench.

\begin{table}[t]
\centering
\small
\caption{Effect of memory size on ChestAgentBench~\cite{fallahpour2025medrax} under two memory-construction settings.}
\label{tab:memory_scale}
\setlength{\tabcolsep}{4pt}
\begin{tabular*}{\linewidth}{@{\extracolsep{\fill}}lccc@{}}
\toprule
Method & $\rho$ & Hit rate (\%) & Overall Acc. \\
\midrule
\multicolumn{4}{l}{\textit{Test-time}} \\
w/o Memory            & 0.0  & 0.0  & 64.0 \\
w/ Memory             & 0.25 & 24.0 & 64.5 \\
w/ Memory             & 0.5  & 30.6 & 64.8 \\
w/ Memory             & 0.75 & 32.1 & 65.0 \\
w/ Memory             & 1.0  & 33.1 & \textbf{65.2} \\
w/ Memory (shuffled)  & 1.0  & $33.1{\pm}0.1$ & $65.2{\pm}0.2$ \\
\midrule
\multicolumn{4}{l}{\textit{Pre-construction (5-fold)}} \\
w/o Memory            & ---  & 0.0  & $64.0{\pm}1.8$ \\
w/ Memory             & ---  & $38.1{\pm}0.1$ & $65.0{\pm}1.6$ \\
\bottomrule
\end{tabular*}\par\vspace{2pt}
{\raggedright\scriptsize Memory-construction settings and the coverage ratio $\rho$ are defined in Section~\ref{sec:setup}. Hit rate: fraction of cases that retrieve at least one memory entry. Shuffled: mean$\pm$std over three random processing orders. Pre-construction scores are mean$\pm$std over $5$-fold cross-validation; per-fold accuracy is reported in Appendix~\ref{sec:cv_results}. \textbf{Bold}: best.\par}
\end{table}

\textbf{Memory Size.}
Table~\ref{tab:memory_scale} (test-time block) reports accuracy as a function of the memory coverage ratio $\rho$. Performance increases monotonically with $\rho$, with $\rho{=}1.0$ improving over the no-memory baseline by $+1.2\%$ overall. This trend indicates that accumulated tool-use experience leads to more reliable tool selection. At low $\rho$, retrieved references are often sparse or absent, so the agent defaults to prompt-driven reasoning. As memory grows, more queries retrieve similar past cases, enabling tool selection based on empirical evidence. Repeating the $\rho{=}1.0$ run with three random processing orders yields $65.2\%{\pm}0.2\%$, indicating that the gain is insensitive to the evaluation order.

\textbf{Test-Time \textit{vs.} Pre-Construction.}
The 5-fold cross-validation result ($65.0\%{\pm}1.6\%$) is close to the test-time result at $\rho{=}1.0$ ($65.2\%$). Under the same five folds, memory improves mean accuracy from $64.0\%$ to $65.0\%$ while slightly reducing the fold-to-fold standard deviation from $1.8$ to $1.6$ percentage points, suggesting a modest improvement in stability across these folds alongside better performance.

\subsection{Cross-Dataset Results}
\label{sec:generalization}

To assess whether the memory mechanism generalizes beyond ChestAgentBench, we evaluate on three additional chest X-ray benchmarks: closed-ended VQA (SLAKE-VQA and CheXbench) and free-form report generation (MIMIC-CXR). Table~\ref{tab:generalization} summarizes the results, where the memory is constructed under the test-time setting with $\rho{=}1.0$. The proposed method consistently improves the baseline across all benchmarks, increasing overall accuracy from $66.2\%$ to $69.7\%$, MF1-5 from $47.9\%$ to $49.1\%$, and recall from $80.6\%$ to $88.5\%$, demonstrating effective generalization across tasks. Conversely, a cross-domain memory constructed only from these three datasets also improves the accuracy on ChestAgentBench from $64.0\%$ to $64.7\%$ (Appendix~\ref{sec:memory_generalization}), indicating that tool-use experience transfers across data sources.

\begin{table*}[t]
\centering
\small
\caption{Cross-dataset generalization of memory-augmented inference on three datasets.}
\label{tab:generalization}
\setlength{\tabcolsep}{1.5pt}
\begin{tabularx}{\linewidth}{@{}l|>{\hsize=0.943\hsize}Y>{\hsize=1.30\hsize}Y>{\hsize=0.943\hsize}Y>{\hsize=0.943\hsize}Y|>{\hsize=0.943\hsize}Y>{\hsize=0.943\hsize}Y>{\hsize=0.943\hsize}Y>{\hsize=0.943\hsize}Y|>{\hsize=1.05\hsize}Y>{\hsize=1.05\hsize}Y@{}}
\toprule
\multirow{2}{*}[-0.5ex]{Method} & \multicolumn{4}{c|}{CheXbench~\cite{chen2024chexagent} (Acc.\,\%)} & \multicolumn{4}{c|}{MIMIC-CXR~\cite{johnson2019mimic} (F1\,\%)} & \multicolumn{2}{c}{SLAKE-VQA~\cite{liu2021slake} (\%)} \\
\cmidrule(r){2-5} \cmidrule(r){6-9} \cmidrule(r){10-11}
& OpenI & Rad-Restruct & SLAKE & Overall & mF1-14 & MF1-14 & mF1-5 & MF1-5 & Acc. & Recall \\
\midrule
w/o Memory & 53.9 & 77.4 & 93.5 & 66.2 & 76.6 & 32.8 & 68.6 & 47.9 & 79.8 & 80.6 \\
w/ Memory               & 59.2 & 78.3 & 94.3 & 69.7 & 77.3   & 32.9 & 69.4   & 49.1   & 87.7 & 88.5 \\
\bottomrule
\end{tabularx}\par\vspace{2pt}
{\raggedright\scriptsize Metrics on MIMIC-CXR and SLAKE-VQA follow the M4CXR protocol~\citep{park2025m4cxr}.\par}
\end{table*}

\subsection{Tool-Use Behaviors}
\label{sec:tool_use_analysis}

To analyze how memory reshapes tool-use behavior, we compare tool trajectories generated by our method and the baseline across the same $2{,}500$ ChestAgentBench cases. We examine two aspects of tool selection: (i) which tools are invoked and their informativeness, and (ii) the order of tool calls and their transitions. Throughout, we refer to the six tools by their abbreviations (Table~\ref{tab:tools}).

\begin{table}[t]
\centering
\small
\setlength{\tabcolsep}{2pt}
\caption{Effect of memory on tool-call frequency and per-call quality on ChestAgentBench (Qwen3-VL 32B).}
\label{tab:frequency_and_score_comparison}
\begin{tabular*}{\linewidth}{@{\extracolsep{\fill}}llrrrrrr@{}}
\toprule
Metric & Method & \texttt{Grd} & \texttt{Cls} & \texttt{Dgn} & \texttt{Rpt} & \texttt{Seg} & \texttt{Vqa} \\
\midrule
Freq. & w/o Mem & 3.36 & 2.54 & 2.54 & 2.06 & 0.76 & 0.12 \\
Freq. & w/ Mem  & 3.44 & 2.40 & 2.47 & 1.78 & 0.71 & 0.10 \\
\midrule
Score     & w/o Mem & 0.679 & 0.417 & 0.145 & 0.169 & 0.474 & 0.446 \\
Score     & w/ Mem  & 0.706 & 0.438 & 0.159 & 0.187 & 0.483 & 0.570 \\
\bottomrule
\end{tabular*}\par\vspace{2pt}
{\raggedright\scriptsize Frequency: average number of calls per case. Score: average LLM-as-a-Judge tool-quality (Eq.~\ref{equation_score}) on common case-tool pairs that occur in both conditions.\par}
\end{table}

\textbf{Frequency and Quality.}
Table~\ref{tab:frequency_and_score_comparison} summarizes the effect of memory on both tool-call frequency and per-call quality. Memory reduces the frequency of most specialized tools, with the largest decreases for case-reusable tools such as \texttt{Rpt} (from $2.06$ to $1.78$ calls per case) and \texttt{Cls} (from $2.54$ to $2.40$). The only increase occurs for \texttt{Grd}, which rises from $3.36$ to $3.44$, consistent with its role as a query-specific localization module whose outputs are not reusable across cases.
For fairness, we compare per-call quality on common case–tool pairs using the LLM-as-a-Judge scores defined in Section~\ref{sec:eval}. Memory improves the average score of all tools, with notable gains for \texttt{Vqa} (from $0.446$ to $0.570$) and \texttt{Rpt} (from $0.169$ to $0.187$). Aggregated tool-level scores in Appendix~\ref{sec:tool_scores_app} show the same trend.
Overall, memory leads to fewer low-value calls and higher-quality retained calls, supporting reliability-aware tool selection. Accumulated experience allows the agent to estimate tool reliability for the current case, suppress less useful calls, and focus on those most likely to improve the final answer.

\begin{figure*}[t]
\centering
\begin{minipage}[c]{0.68\textwidth}
\centering
\includegraphics[width=\linewidth]{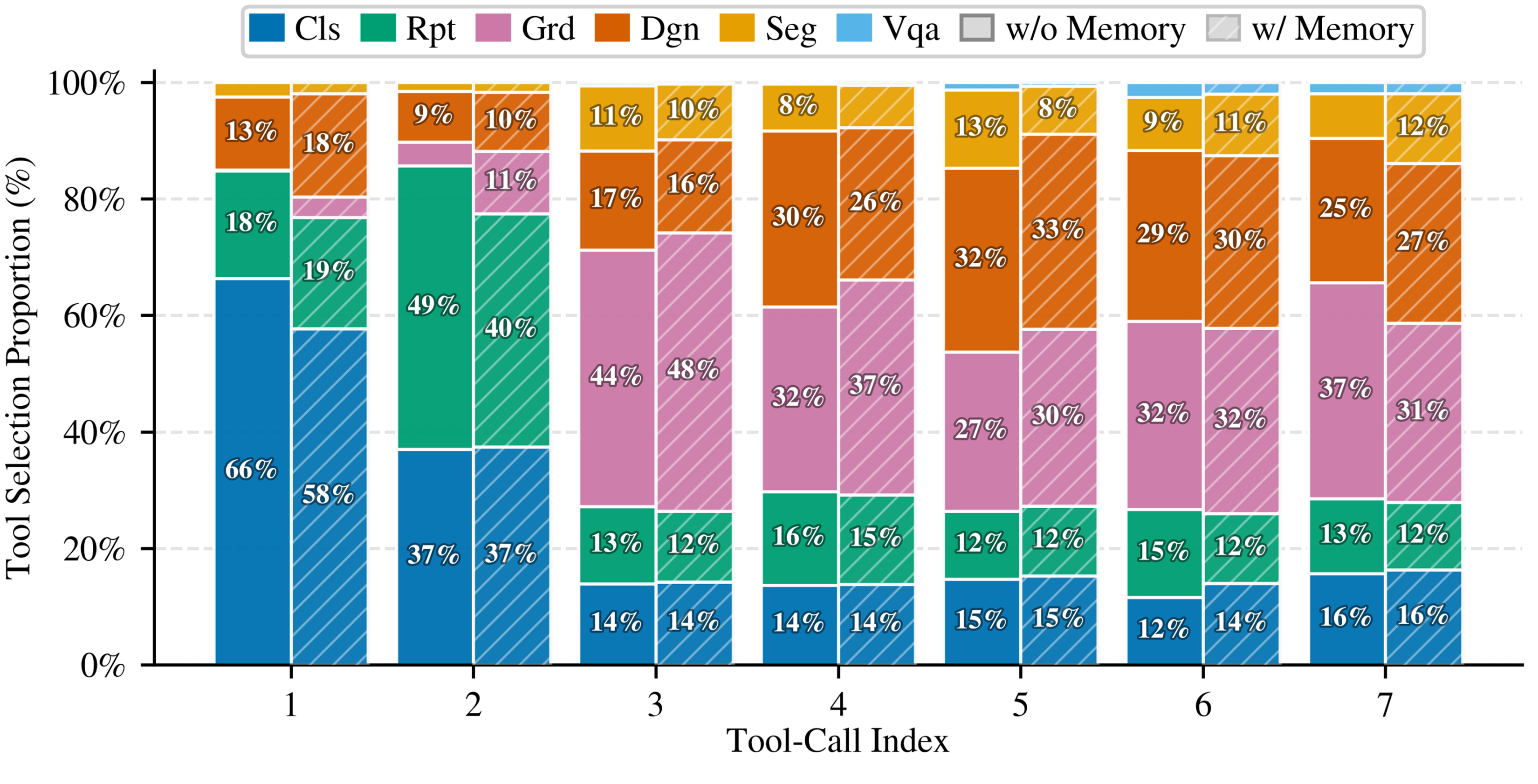}
\end{minipage}\hfill
\begin{minipage}[c]{0.305\textwidth}
\caption{\textbf{Tool selection dynamics across tool-call steps.} Per-step usage proportion of each tool under Baseline and Ours (with memory). Memory accelerates the convergence to the final tool mix: low-final-proportion tools (\texttt{Cls}, \texttt{Rpt}) are suppressed earlier (steps 1--2), while high-final-proportion tools (\texttt{Grd}, \texttt{Dgn}) are amplified within the same window. From step 4 onwards, the per-step distributions of the two conditions are within $5\%$ of each other.}
\label{fig:tool_dynamics}
\end{minipage}
\end{figure*}

\textbf{Tool Order.}
Beyond aggregated frequency, memory also reshapes the temporal arrangement of tool calls. Figure~\ref{fig:tool_dynamics} shows that memory accelerates the trajectory's convergence to its final tool mix: low-final-proportion tools are suppressed earlier (\texttt{Cls} drops $8\%$ at step 1 and \texttt{Rpt} drops $9\%$ at step 2) while high-final-proportion tools (\texttt{Grd} and \texttt{Dgn}) are amplified within the same window; from step 4 onwards, per-step shifts are below $5\%$, indicating that memory mainly accelerates the early reasoning phase rather than altering the late-stage tool mix. A complementary analysis of the inter-tool transition matrix $P(T_{j+1}{\mid}T_j)$ (Appendix~\ref{sec:transition_app}) further shows that the dominant shift concentrates on the outgoing distribution of \texttt{Vqa}: probability mass moves from the cross-tool path \texttt{Vqa}$\to$\texttt{Cls} ($-44\%$ relative) to the self-continuation \texttt{Vqa}$\to$\texttt{Vqa} ($+243\%$ relative), all other transitions changing by less than $0.05$ in absolute terms. This replaces a case-independent verification loop with a reliability-aware within-tool continuation.

\textbf{Vocabulary Entropy.}
As a control, both per-call and corpus-level word-level Shannon entropy of prompts and responses show changes that are mostly concentrated below $0.10$ bits, with the main exception being corpus-level Dgn responses. This suggests that memory does not improve performance primarily by increasing verbosity or lexical diversity (see Appendix~\ref{sec:entropy_analysis}). Appendix~\ref{sec:tool_arg_examples_app} further provides qualitative examples showing that memory can modify tool arguments, such as focusing prompts on clinically relevant cues or correcting image inputs for multi-view comparison.

\paragraph{Summary.}
The above views converge on a concise statement: \emph{memory's effect on tool use reduces to reliability-aware tool selection.} Conditioned on empirical evidence retrieved from similar past cases, the agent (i) avoids invocations that prior experience marks as unlikely to help on the current case---lowering the frequency of redundant tools while raising the per-call usefulness of those that survive; and (ii) advances the most reliable tools earlier in the trajectory and replaces case-independent cross-tool verification with within-tool deepening. Both effects share the same underlying mechanism---substituting prompt-only tool arbitration with case-specific reliability evidence---and together produce the accuracy gains reported in Section~\ref{sec:main_results}.

\subsection{Hyperparameters}
\label{sec:hp_ablation}

\begin{table}[t]
\centering
\small
\caption{Effect of retrieval hyperparameters.}
\label{tab:hyperparameters}
\setlength{\tabcolsep}{5pt}
\begin{tabularx}{\linewidth}{@{}c|YYY|YY@{}}
\toprule
$k$ & 3& 3&3&1&10\\
\midrule
$\tau$     & 0.9&0.99&0.999 &0.99&0.99        \\
\midrule
Overall Acc.&64.8&\textbf{65.2}&65.0 &64.8&64.3 \\
\bottomrule
\end{tabularx}\\[2pt]
\end{table}
We study the sensitivity of memory-augmented inference to the retrieval hyperparameters $k$ and $\tau$ on ChestAgentBench using Qwen3-VL 32B. As shown in Table~\ref{tab:hyperparameters}, we vary one hyperparameter at a time around the default setting $(k{=}3,\tau{=}0.99)$. When fixing $k{=}3$, the default threshold $\tau{=}0.99$ achieves the best accuracy. A lower threshold ($\tau{=}0.9$) retrieves less relevant historical cases, while a higher threshold ($\tau{=}0.999$) reduces useful memory coverage. When fixing $\tau{=}0.99$, increasing $k$ from $1$ to $3$ improves performance, indicating that multiple similar cases provide richer tool-reliability references. However, further increasing $k$ to $10$ degrades performance, likely because it includes fewer relevant cases, diluting the retrieved evidence. Overall, $(k{=}3,\tau{=}0.99)$ provides the best trade-off between relevance and coverage. A complementary ablation on tag-based feature extraction is provided in Appendix~\ref{sec:ablation_tag_extraction}.

\vspace{3pt}
\section{Conclusion}
\label{sec:conclusion}
\vspace{2pt}
We present EATS, an experience-augmented strategy for improving tool selection in medical reasoning agents. By constructing a long-term memory of tool-use trajectories and retrieving relevant past experiences at inference time, EATS enables reliability-aware decision-making without additional training or supervision. This design effectively addresses the challenge of uncertain and case-dependent neural tool outputs in clinical workflows. Extensive experiments across four benchmarks and three multimodal foundation models demonstrate consistent improvements in both tool selection accuracy and downstream reasoning tasks, including VQA and report generation. These results highlight the value of leveraging accumulated experience to enable scalable, adaptive medical tool use.


\clearpage
\section*{LLM Usage Statement}

The authors used GPT-6.1 Sol to assist with language polishing of author-written text, including grammar and spelling checks, to improve clarity and readability for non-native English writers. The tool also assisted in editing and refining author-produced code and implementation, as well as in refining author-developed experimental designs, supporting experiment execution, and assisting with the analysis of results. Its role in these activities was limited to assistance with author-produced material. Neither the full manuscript nor major portions of it were submitted to an LLM provider for reviewer-style feedback. The authors take full responsibility for all submitted content.

{
    \small
    \bibliographystyle{ieeenat_fullname}
    \bibliography{main}
}

\clearpage
\maketitlesupplementary
\appendix

The appendix provides supplementary material along seven aspects. First, it expands the tool-use behavior analysis with the full inter-tool transition matrix in Appendix~\ref{sec:transition_app}, qualitative examples of memory-induced changes to tool arguments in Appendix~\ref{sec:tool_arg_examples_app}, and aggregated tool-level LLM-as-a-Judge scores in Appendix~\ref{sec:tool_scores_app}. Second, it reports additional experimental details and robustness checks, including per-fold cross-validation results in Appendix~\ref{sec:cv_results}, the tag-based retrieval ablation in Appendix~\ref{sec:ablation_tag_extraction}, the validation of the label-free judge in Appendix~\ref{sec:judge_validation}, and the generalization of memory across datasets in Appendix~\ref{sec:memory_generalization}. Third, it documents the prompt used to extract structured question tags for memory retrieval in Appendix~\ref{sec:prompt_extract}. Fourth, it specifies the complete tool-contribution evaluation prompt and scoring rubrics in Appendix~\ref{sec:prompt_eval}. Fifth, it provides the full per-call and corpus-level entropy analysis that complements the tool-use study in Appendix~\ref{sec:entropy_analysis}. Sixth, it reports the computational cost of EATS in Appendix~\ref{sec:overhead} and discusses the main empirical scope and cross-domain generalization limitations in Appendix~\ref{sec:limitations}. Finally, it documents the MedRAX system prompt and default tool interface used in our main experiments in Appendix~\ref{sec:implementation_details_app}.

\section{Inter-Tool Transition Matrix}
\label{sec:transition_app}

\begin{figure*}[t]
\centering
\includegraphics[width=\textwidth]{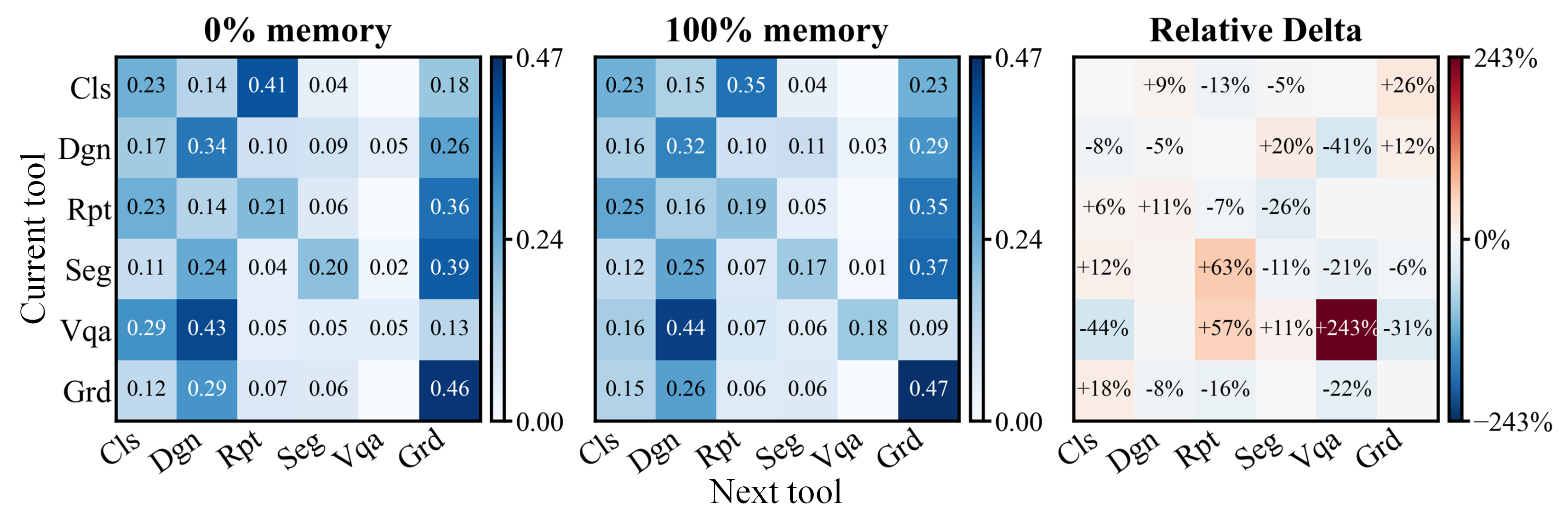}
\caption{\textbf{Tool-transition matrices: memory replaces cross-tool verification with within-tool continuation.} Left and middle: the tool-transition matrix $P(T_{j+1}{\mid}T_j)$ for Baseline and Memory respectively (rows: current tool, columns: next tool). Right: \emph{relative change} of each transition probability, defined as $(P_{\mathrm{Mem}}{-}P_{\mathrm{Base}})/P_{\mathrm{Base}}$. The dominant change concentrates on the \texttt{Vqa} row: probability mass moves from the cross-tool path \texttt{Vqa}$\to$\texttt{Cls} ($-44\%$ relative) to the self-continuation \texttt{Vqa}$\to$\texttt{Vqa} ($+243\%$ relative); all other transitions shift by less than $0.05$ in absolute terms.}
\label{fig:tool_transition_app}
\end{figure*}

Figure~\ref{fig:tool_transition_app} presents the full inter-tool transition matrix referenced in Section~\ref{sec:tool_use_analysis}. We define the \emph{tool-transition matrix} $P(T_{j+1}{\mid}T_j)$ as the empirical conditional probability that tool $T_{j+1}$ is invoked at step $j{+}1$ given that tool $T_j$ was invoked at step $j$; each cell is therefore the frequency of one tool following another in the agent's trajectory. The dominant change concentrates on the outgoing distribution of \texttt{Vqa}: the \texttt{Vqa}$\to$\texttt{Cls} probability drops from $0.29$ to $0.16$ (an absolute decrease of $0.13$, $-44\%$ relative change), while the self-transition \texttt{Vqa}$\to$\texttt{Vqa} rises symmetrically from $0.05$ to $0.18$ (an absolute increase of $0.12$, $+243\%$ relative change); all other transitions shift by less than $0.05$ in absolute terms. The Baseline pattern---querying \texttt{Vqa} and then switching to \texttt{Cls} to confirm---is a case-independent verification loop. With per-case reliability evidence available in $\Gamma(\mathcal{X})$, the agent deepens \texttt{Vqa}-based probing instead of switching to a different tool, producing a clear path-level pattern of reliability-aware tool selection.

\section{Per-Fold Cross-Validation Results}
\label{sec:cv_results}

Table~\ref{tab:cv_folds} reports the per-fold accuracy of the memory-free MedRAX baseline and our method on
ChestAgentBench under 5-fold cross-validation.
Each per-fold accuracy is averaged over three runs.

\begin{table*}[t]
\centering
\small
\caption{Per-fold accuracy (\%) of the memory-free baseline and our method (Qwen3-VL 32B) on ChestAgentBench
  under 5-fold cross-validation.}
\label{tab:cv_folds}
\setlength{\tabcolsep}{5pt}
\begin{tabular*}{\linewidth}{@{\extracolsep{\fill}}lcccccccc@{}}
\toprule
Fold & Overall & Detect. & Classif. & Local. & Compar. & Relat. & Diagn. & Charact. \\
\midrule
\multicolumn{9}{l}{\textit{w/o Memory (MedRAX)}} \\
1    & 65.1  & 65.1 & 64.7 & 69.5 & 57.3 & 59.8 & 62.5 & 65.9 \\
2    & 64.5  & 63.8 & 66.6 & 63.4 & 61.2 & 61.2 & 65.2 & 67.5 \\
3    & 61.3  & 67.9 & 62.5 & 61.8 & 63.8 & 65.8 & 61.4 & 68.6 \\
4    & 63.3  & 66.7 & 60.6 & 66.5 & 61.6 & 64.6 & 66.0 & 63.1 \\
5    & 65.9  & 65.0 & 62.6 & 64.3 & 65.6 & 66.6 & 59.9 & 66.4 \\
\midrule
Mean & 64.0  & 65.7 & 63.4 & 65.1 & 61.9 & 63.6 & 63.0 & 66.3 \\
Std & $\pm 1.8$  & $\pm 1.6$ & $\pm 2.3$ & $\pm 3.0$ & $\pm 3.1$ & $\pm 3.0$ & $\pm 2.6$ & $\pm 2.1$ \\
\midrule
\multicolumn{9}{l}{\textit{w/ Memory (EATS)}} \\
1    & 66.6       & 67.8       & 65.5       & 68.2       & 65.8       & 64.0       & 66.5       & 68.9       \\
2    & 62.9       & 65.5       & 62.9       & 62.9       & 59.8       & 62.8       & 61.6       & 64.8       \\
3    & 64.6       & 65.6       & 61.6       & 69.2       & 62.7       & 66.5       & 61.1       & 67.7       \\
4    & 66.7       & 68.9       & 67.5       & 65.5       & 66.5       & 68.7       & 65.4       & 64.3       \\
5    & 64.3       & 68.8       & 65.1       & 63.2       & 60.6       & 60.8       & 64.5       & 67.2       \\
\midrule
Mean & 65.0       & 67.3       & 64.5       & 65.8       & 63.1       & 64.6       & 63.8       & 66.6       \\
Std  & $\pm 1.6$   & $\pm 1.7$  & $\pm 2.3$  & $\pm 2.9$  & $\pm 3.0$  & $\pm 3.1$  & $\pm 2.4$  & $\pm 2.0$  \\
\bottomrule
\end{tabular*}
\end{table*}

\section{Ablation on Tag-Based Feature Extraction}
\label{sec:ablation_tag_extraction}

We further ablate the construction of the retrieval key used by long-term
memory. The default variant first applies the tag extraction prompt
$P_{\mathrm{ext}}$ to the question and then embeds the resulting structured tag
string, i.e., $\mathbf{c}=f_{\theta}(Q;P_{\mathrm{ext}})$ and
$\mathbf{z}=\phi(\mathbf{c})$. The ablated variant removes this prompt and
directly embeds the raw question, i.e., $\mathbf{z}=\phi(Q)$, while keeping the
same embedding model, memory, and retrieval hyperparameters
$(k{=}3,\tau{=}0.99)$. The extracted tags contain six semantic fields:
question type, symptoms, pathological findings, demographic information, risk
factors, and explicitly negated information.

\begin{table}[t]
\centering
\small
\caption{Ablation of the tag extraction prompt for memory retrieval on
ChestAgentBench with Qwen3-VL 32B. Both variants use the same embedding model
$\phi(\cdot)$ and the same retrieval setting $(k{=}3,\tau{=}0.99)$.}
\label{tab:tag_extraction_ablation}
\setlength{\tabcolsep}{3pt}
\begin{tabular*}{\linewidth}{@{\extracolsep{\fill}}llc@{}}
\toprule
Variant & Embedded Text Form & Overall Acc. (\%) \\
\midrule
w/o $P_{\mathrm{ext}}$ & Raw question $Q$                         &$64.7$ \\
w/ $P_{\mathrm{ext}}$  & Structured tags $f_{\theta}(Q;P_{\mathrm{ext}})$ & \textbf{65.2} \\
\bottomrule
\end{tabular*}
\end{table}

Table~\ref{tab:tag_extraction_ablation} shows that the improvement of performance does not come from simply applying an embedding model to the question text. Using $P_{\mathrm{ext}}$ converts the question into a compact, clinically aligned representation before embedding, which makes retrieval more sensitive to the attributes that determine whether historical tool-use experience is transferable. In contrast, the w/o $P_{\mathrm{ext}}$ variant mixes task phrasing, irrelevant surface words, and clinical concepts in the same raw-question embedding space, leading to less relevant retrieved memory entries and a lower overall accuracy than the default structured-tag setting.

\section{Validation of the Label-Free Judge}
\label{sec:judge_validation}

EATS scores each tool event with the same foundation model $f_{\theta}$ and without ground-truth labels (Section~\ref{sec:eval}). To examine whether this label-free scoring limits the usefulness of the memory, we replace it with a GT-based judge that scores each tool event according to whether its output helps the agent reach the ground-truth answer, with scores ranging from $-1$ (harmful) to $1$ (helpful), while keeping memory construction, retrieval, and inference unchanged. Table~\ref{tab:judge_validation} shows that the two judges yield nearly identical accuracy ($65.2\%$ \vs $65.1\%$). Label-free self-evaluation is therefore on par with ground-truth supervision for constructing useful memory, which supports the ability of EATS to evolve its memory without labels.

\begin{table}[t]
\centering
\small
\caption{Validation of the label-free judge on ChestAgentBench (Qwen3-VL 32B, test-time setting, $\rho{=}1.0$).}
\label{tab:judge_validation}
\setlength{\tabcolsep}{3pt}
\begin{tabular*}{\linewidth}{@{\extracolsep{\fill}}lcc@{}}
\toprule
Judge & Ground-truth labels & Overall Acc. (\%) \\
\midrule
Self-judge (default) & No  & 65.2 \\
GT-based judge       & Yes & 65.1 \\
\bottomrule
\end{tabular*}
\end{table}

\section{Generalization of Memory across Datasets}
\label{sec:memory_generalization}

To test whether tool-use experience transfers beyond the dataset from which it is collected, we compare two memory sources on ChestAgentBench with Qwen3-VL 32B: a same-domain memory pre-constructed from the other four folds of ChestAgentBench (Section~\ref{sec:setup}), and a cross-domain memory constructed only from CheXbench, MIMIC-CXR, and SLAKE-VQA. As shown in Table~\ref{tab:memory_generalization}, both memories improve over the memory-free baseline ($64.0\%$). The cross-domain memory yields a smaller gain ($+0.7$) than the same-domain memory ($+1.0$), consistent with the larger distribution shift between data sources, but the gain does not vanish.

\begin{table}[t]
\centering
\small
\caption{Generalization validation of memory on same- and cross-domain datasets. All variants are evaluated on ChestAgentBench with Qwen3-VL 32B.}
\label{tab:memory_generalization}
\setlength{\tabcolsep}{3pt}
\begin{tabularx}{\linewidth}{@{}l>{\raggedright\arraybackslash}Xcc@{}}
\toprule
Memory & Source datasets & Overall Acc. & $\Delta$ \\
\midrule
w/o Memory   & ---                               & 64.0           & ---    \\
Same-domain  & ChestAgentBench (other four folds) & $65.0{\pm}1.6$ & $+1.0$ \\
Cross-domain & CheXbench, MIMIC-CXR, SLAKE-VQA    & 64.7           & $+0.7$ \\
\bottomrule
\end{tabularx}
\end{table}

\section{Qualitative Examples of Memory-Induced Argument Changes}
\label{sec:tool_arg_examples_app}

Beyond the population-level effects on tool selection reported in Section~\ref{sec:tool_use_analysis}, memory also influences the per-call arguments---prompts and image paths---supplied to each tool invocation. Table~\ref{tab:tool_arg_examples_app} shows two representative cases for the \texttt{Dgn} tool. In the first case, the baseline prompt is a broad multi-pathology query for which the tool returns garbled output ($s{=}{-}1.00$); under memory, the retrieved past-case experience steers the agent toward the relevant cue (foreign-body during intubation), yielding a focused prompt that the tool answers correctly ($s{=}{+}1.00$). In the second case, the baseline supplies a duplicated single image where the question requires multi-view comparison; memory steers the agent toward the correct two-image input, and the tool produces a usable cross-view interpretation ($\Delta s{=}{+}1.27$). These are anecdotal examples and we do not claim a population-level effect on argument quality.

\begin{table*}[t]
\centering
\small
\setlength{\tabcolsep}{3pt}
\caption{Examples showing how memory changes the arguments (prompt and image inputs) of neural tool calls, which in turn changes tool effectiveness. Both examples are from the \texttt{Dgn} tool but on different cases; the question identifier (QID) is included in the original-question column.}
\label{tab:tool_arg_examples_app}
\begin{tabularx}{\linewidth}{@{}p{0.45in} X X X p{0.42in}@{}}
\hline
Factor & Original question & No Memory & Memory & $\Delta$ score \\
\hline
Prompt
& \texttt{QID 11340\_8353800751074}.
Based on day 1 and day 4 chest X-rays in an 85-year-old man with respiratory failure and difficult tracheal intubation, what most likely caused the obstruction during intubation?
& the prompt broadly asked about airway obstruction, pleural effusion, tracheal deviation, and lung volume loss. The output was garbled and unusable. Score: \texttt{-1.00}.
& the prompt explicitly asked about dental prostheses, foreign bodies, and obstructing structures during intubation. This focused the tool on the right cue and it was directly cited by the agent. Score: \texttt{1.00}.
& \texttt{+2.00} \\
\hline
Image paths
& \texttt{QID 3561\_8353801511489}. Based on the chest X-rays provided, what is the most likely cause for the rounded opacity near the left costophrenic angle observed in Figure 1 and confirmed on the lateral view in Figure 5?
& \texttt{image\_paths} contained the same image (\texttt{figure\_1.jpg}) instead of a true multi-view pair, limiting its ability to compare the PA and lateral findings. The resulting evidence was weak and partially unhelpful. Score: \texttt{-0.35}.
& \texttt{image\_paths} contained the correct two-image input (\texttt{figure\_1.jpg} and \texttt{figure\_5.jpg}), matching the original task setup. This enabled a proper cross-view interpretation and substantially improved the usefulness of the output. Score: \texttt{0.92}.
& \texttt{+1.27} \\
\hline
\end{tabularx}
\end{table*}

\section{Aggregated Tool-Level LLM-as-a-Judge Scores}
\label{sec:tool_scores_app}

We report this aggregated view for completing the conclusion---memory improves the tool quality---in Section~\ref{sec:tool_use_analysis} rely on the paired view.
Table~\ref{tab:tool_scores_app} reports the aggregated tool-level LLM-as-a-Judge scores referenced in Section~\ref{sec:tool_use_analysis}. Unlike the paired view in Table~\ref{tab:frequency_and_score_comparison}---which show the effect of memory on the common case-tool pairs---this aggregated view averages each criterion over \emph{all} calls of each tool, regardless of which case the call belongs to. The results from the two views are different because the set of cases on which a tool is invoked differs between conditions; but the score in the aggregated view consistently reflects that our memory improves the overall tool quality.

\begin{table*}[t]
\centering
\small
\caption{Aggregated tool-level LLM-as-a-Judge scores under no-memory and memory-augmented test-time inference (Qwen3-VL 32B). Each cell is averaged over all calls of the tool across all cases; \texttt{Score} is the mean of the five evaluation criteria defined in Section~\ref{sec:eval}.}
\label{tab:tool_scores_app}
\setlength{\tabcolsep}{5.5pt}
\begin{tabular*}{\linewidth}{@{\extracolsep{\fill}}lrrrrrr@{}}
\toprule
Method & Relevance & Increment & Diagnostic & Citation & Consistency & Score \\
\midrule
\texttt{Grd} (w/o Memory)  &  0.781 &  0.698 &  0.491 &  0.734 &  0.529 &  0.646 \\
\texttt{Grd} (w/ Memory)     &  0.782 &  0.711 &  0.504 &  0.736 &  0.540 &  0.655 \\
\texttt{Dgn} (w/o Memory) & 0.236 & 0.137 & -0.343 & 0.335 & -0.039 & 0.065 \\
\texttt{Dgn} (w/ Memory) & 0.385 & 0.132 & -0.131 & 0.305 & -0.002 & 0.138 \\
\texttt{Cls} (w/o Memory)  &  0.595 &  0.596 &  0.190 &  0.337 &  0.225 &  0.389 \\
\texttt{Cls} (w/ Memory)     &  0.613 &  0.619 &  0.215 &  0.380 &  0.252 &  0.416 \\
\texttt{Rpt} (w/o Memory)  &  0.353 &  0.180 & -0.123 &  0.292 & -0.129 &  0.115 \\
\texttt{Rpt} (w/ Memory)     &  0.365 &  0.220 & -0.094 &  0.300 & -0.095 &  0.139 \\
\texttt{Seg} (w/o Memory)  &  0.468 &  0.452 &  0.121 &  0.322 &  0.220 &  0.317 \\
\texttt{Seg} (w/ Memory)     &  0.452 &  0.441 &  0.121 &  0.354 &  0.225 &  0.318 \\
\texttt{Vqa} (w/o Memory)  &  0.716 &  0.629 &  0.148 &  0.577 &  0.117 &  0.437 \\
\texttt{Vqa} (w/ Memory)     &  0.705 &  0.659 &  0.275 &  0.606 &  0.133 &  0.476 \\
\bottomrule
\end{tabular*}
\end{table*}

\section{Question Tags Extraction Prompt}
\label{sec:prompt_extract}

This appendix reports the exact prompt $P_{\mathrm{ext}}$ used to extract structured question tags of ChestAgentBench~\citep{fallahpour2025medrax} for constructing the memory retrieval key.

\begin{quote}
\textbf{Role and task.}
You are a medical tagging assistant for chest imaging MCQ questions. Extract tags from the question stem and options as JSON with the exact schema below.

\textbf{Output schema.}
The JSON object must contain the following fields:
\begin{itemize}[leftmargin=1.5em,itemsep=0.1em]
    \item \emph{question\_type}: a list of question-type labels.
    \item \emph{symptoms}: a list of symptom tags.
    \item \emph{demographics}: a list of demographic tags.
    \item \emph{risk\_factors}: a list of risk-factor tags.
    \item \emph{pathological\_findings}: a list of canonical pathology labels.
    \item \emph{denied\_information}: an object containing three lists:
    \emph{symptoms}, \emph{pathological\_findings}, and \emph{risk\_factors}.
\end{itemize}

\textbf{Question type labels.}
The field \emph{question\_type} must be a subset of exactly the following seven
labels, using these exact names:
\begin{itemize}[leftmargin=1.5em,itemsep=0.1em]
    \item \textbf{Detection}: Identifying specific findings, e.g., whether a
    nodule is present in the right upper lobe.
    \item \textbf{Classification}: Classifying specific findings, e.g., whether
    a mass appears benign or malignant.
    \item \textbf{Localization}: Precise positioning of findings, e.g., the
    bronchopulmonary segment containing a mass.
    \item \textbf{Comparison}: Analyzing relative sizes and positions, e.g.,
    how pleural effusion volume changed compared to prior imaging.
    \item \textbf{Relationship}: Understanding relationships between findings,
    e.g., whether mediastinal lymphadenopathy correlates with a lung mass.
    \item \textbf{Diagnosis}: Interpreting findings for clinical decisions,
    e.g., inferring the likely diagnosis from a CXR.
    \item \textbf{Characterization}: Describing finding attributes, e.g.,
    whether nodule margins are smooth, spiculated, or irregular.
\end{itemize}

\textbf{Extraction rules.}
\begin{enumerate}[leftmargin=1.5em,itemsep=0.1em]
    \item \emph{pathological\_findings} must be selected only from this closed
    set of 20 labels: Mass, Effusion, Pleural Effusion, Consolidation, Nodule,
    Calcification, Pneumothorax, Lymphadenopathy, Pneumonia, Emphysema,
    Interstitial findings, Bronchiectasis, Atelectasis, Fibrosis, Edema,
    Cavitation, Fracture, Tuberculosis, Metastasis, and Cardiomegaly.
    \item Extract \emph{pathological\_findings} from the question stem and
    options only, using text evidence only and no image inference.
    \item If a synonym or variant appears, map it to the closest canonical label
    from the 20 labels above.
    \item Do not output any finding outside this 20-label set; deduplicate the
    output.
    \item Symptoms are subjective, patient-reported experiences indicating
    illness or injury, and are not directly observable by others, e.g., pain,
    nausea, fatigue, dizziness, headache, cough, fever, chest pain, and dyspnea.
    \item Demographics are statistical attributes describing people, such as
    age, sex/gender, pregnancy status, ethnicity, occupation, education, or
    income, when explicitly stated and clinically relevant.
    \item Risk factors are characteristics, conditions, exposures, or behaviors
    that increase adverse outcome likelihood, e.g., smoking, TB exposure,
    immunosuppression, alcohol misuse, occupational dust exposure, and family
    history.
    \item \emph{denied\_information} must contain only explicitly negated items
    from the question stem for symptoms, pathological findings, and risk
    factors, e.g., ``no history of smoking'', ``afebrile'', ``no chest pain'',
    and ``without pleural effusion''. Never infer negation.
    \item \emph{denied\_information.pathological\_findings} must use the same
    20 canonical pathological finding labels.
    \item Keep all lists concise and deduplicated. If no evidence is present for
    a field, return an empty list.
\end{enumerate}

\textbf{Output constraint.}
Return JSON only. No markdown.
\end{quote}

\section{Tool Contribution Evaluation Prompt}
\label{sec:prompt_eval}

The prompt $P_{\mathrm{eval}}$ is issued as a single call to $f_{\theta}$ after
each completed reasoning trajectory.
It receives the question, answer options, all tool calls with their outputs, the
agent's full reasoning trace, and the agent's final answer.
The model is instructed \emph{not} to judge whether the final answer is correct,
but only to assess tool output quality and its use in reasoning.

For each tool call, the model assigns five independent sub-scores on a five-point
scale $\{-1.0,\,-0.5,\,0.0,\,0.5,\,1.0\}$ and provides a one-sentence rationale
per criterion.
The output is returned as structured JSON.
The five criteria and their scoring rubrics are defined as follows.

\paragraph{MOQ-1: Clinical Relevance (\texttt{moq\_relevance}).}
Measures whether the tool output is medically relevant to the specific question.
\begin{quote}
\begin{tabularx}{\linewidth}{@{}l X@{}}
$+1.0$ & Directly targets the pathology or task type asked about. \\
$+0.5$ & Related to the general clinical domain but only partially on-topic. \\
$\phantom{+}0.0$ & Generic or tangential output with no specific relevance. \\
$-0.5$ & Off-topic output that may distract from the actual question. \\
$-1.0$ & Output addresses a completely different condition or task.
\end{tabularx}
\end{quote}

\paragraph{MOQ-2: Informational Increment (\texttt{moq\_increment}).}
Measures whether the tool output adds findings beyond what the question stem
already states.
\begin{quote}
\begin{tabularx}{\linewidth}{@{}l X@{}}
$+1.0$ & Provides concrete new findings (e.g.\ measurements, classifications, anatomical regions) not present in the question stem. \\
$+0.5$ & Adds some new detail but largely overlaps with the stem. \\
$\phantom{+}0.0$ & Restates question content without adding anything new. \\
$-0.5$ & Introduces information that contradicts or confuses the stem. \\
$-1.0$ & Output is empty, garbled, or a clear tool failure.
\end{tabularx}
\end{quote}

\paragraph{MOQ-3: Diagnostic Discriminability (\texttt{moq\_diagnostic}).}
Measures whether the tool output helps narrow the differential diagnosis.
\begin{quote}
\begin{tabularx}{\linewidth}{@{}l X@{}}
$+1.0$ & Clearly supports or excludes one or more plausible diagnoses or options on clinical grounds. \\
$+0.5$ & Suggestive toward a diagnostic direction but not conclusive. \\
$\phantom{+}0.0$ & Neutral---does not distinguish between competing diagnoses. \\
$-0.5$ & Contains a clinically incorrect finding that could push reasoning toward a wrong diagnosis. \\
$-1.0$ & Asserts a finding directly contradicted by image evidence or clinical context.
\end{tabularx}
\end{quote}

\paragraph{RCC-1: Citation Fidelity (\texttt{rcc\_citation}).}
Measures whether the agent's reasoning explicitly references this tool output.
\begin{quote}
\begin{tabularx}{\linewidth}{@{}l X@{}}
$+1.0$ & Agent directly quotes or paraphrases specific values or findings from this output. \\
$+0.5$ & Agent mentions the tool or its result in passing without detailed use. \\
$\phantom{+}0.0$ & Agent does not reference this tool output at all. \\
$-0.5$ & Agent acknowledges the tool but explicitly dismisses or contradicts it. \\
$-1.0$ & Agent misrepresents the tool output (cites wrong values or inverted findings).
\end{tabularx}
\end{quote}

\paragraph{RCC-2: Answer Consistency (\texttt{rcc\_consistency}).}
Measures whether the tool output is logically consistent with the agent's final
answer. The model is instructed not to judge whether the answer is correct, only
whether the tool output logically aligns with the clinical reasoning that led to it.
\begin{quote}
\begin{tabularx}{\linewidth}{@{}l X@{}}
$+1.0$ & Tool output directly supports the clinical reasoning leading to the agent's answer. \\
$+0.5$ & Tool output is compatible with the answer but does not strongly support it.\\
$\phantom{+}0.0$ & Tool output is neutral---consistent with multiple options equally. \\
$-0.5$ & Tool output is more consistent with a different option. \\
$-1.0$ & Tool output directly contradicts the clinical basis of the agent's answer.
\end{tabularx}
\end{quote}

The five scores are aggregated into a single scalar $s_j \in [-1,1]$ via
equal-weight averaging.
The JSON output format requires one \texttt{rationale} field containing five
sentences, one per criterion, to ensure transparent and auditable scoring.


\section{Per-Call and Corpus-Level Entropy Analysis}
\label{sec:entropy_analysis}

This appendix provides the complete per-call and corpus-level entropy analysis summarized in Section~\ref{sec:tool_use_analysis}. Two complementary entropy views are reported: per-call entropy, which characterizes the lexical content of any single tool invocation, and corpus-level cumulative entropy, which characterizes the diversity of the union of all calls of a given tool across the dataset.

\begin{figure*}[t]
\centering
\includegraphics[width=0.8\textwidth]{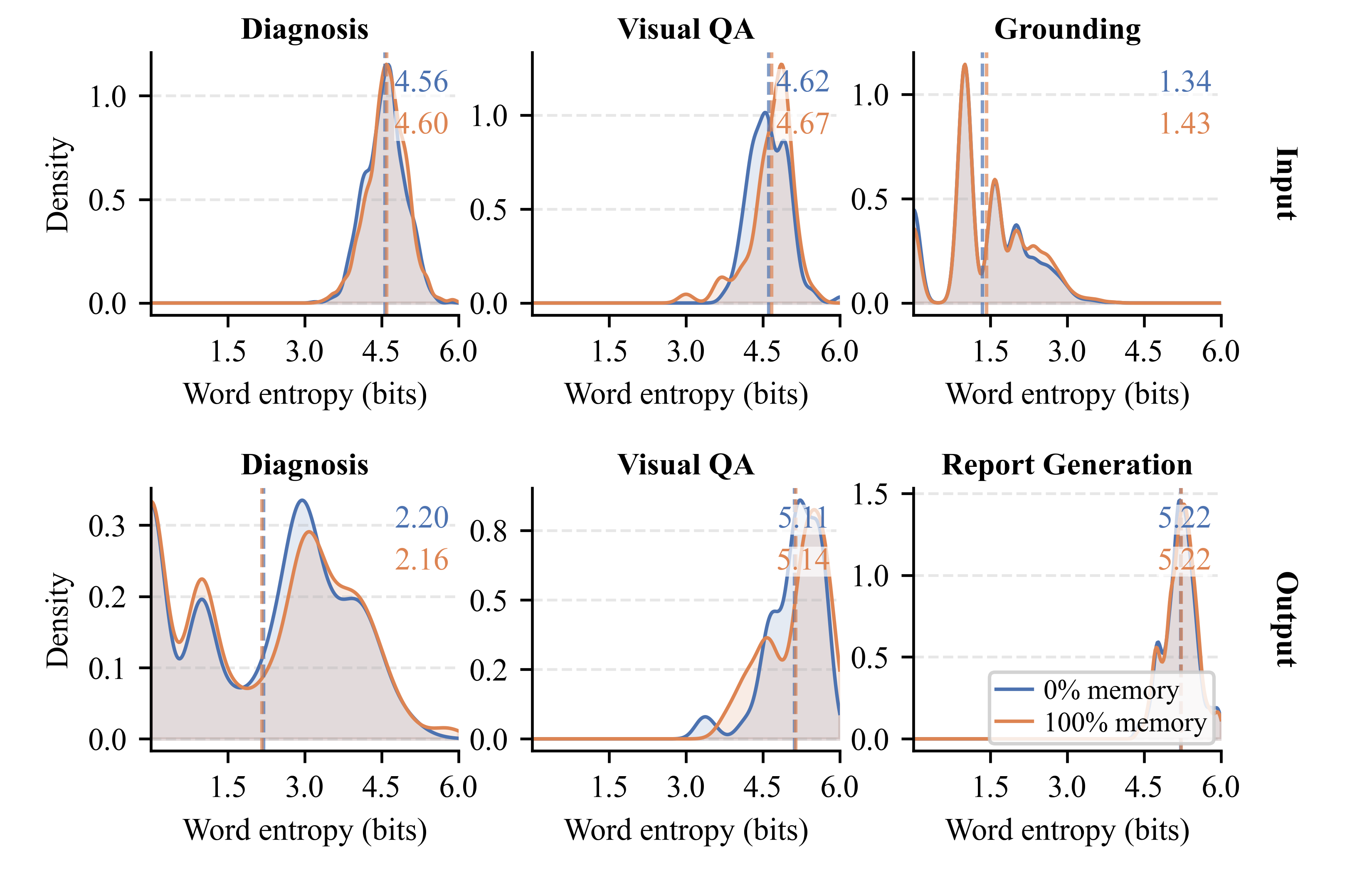}
\caption{\textbf{Per-call input/output entropy: memory does not change single-call vocabulary.}
Kernel density estimates of the word-level Shannon entropy of each individual tool call (top row: inputs; bottom row: outputs); dashed lines mark condition means.
All differences remain within $0.10$ bits, showing that memory's effect does not reside in the lexical content of individual calls.}
\label{fig:info_entropy_app}
\end{figure*}

\paragraph{Per-call entropy (Figure~\ref{fig:info_entropy_app}).}
For each tool call we compute the word-level Shannon entropy $H{=}-\sum_{w\in\mathcal{V}} p(w)\log_2 p(w)$, with $p(w)$ the relative frequency of word $w$ in the text (in bits); higher $H$ indicates richer vocabulary. Examining $H$ for both the prompt and the response of each call, we find that all differences are small ($|\Delta|{<}0.10$ bits) across the analyzed tools. Memory therefore does not substantially change the vocabulary of any single call. This is the expected behavior: specialized tools such as \texttt{Dgn} and \texttt{Rpt} are deterministic given their inputs, so changing the agent's memory state cannot alter what a single invocation returns. The absence of a per-call effect also serves as a useful control---any corpus-level change observed below must stem from a change in \emph{which} calls are issued, not in how any single call is phrased.

\begin{figure*}[t]
\centering
\includegraphics[width=0.8\textwidth]{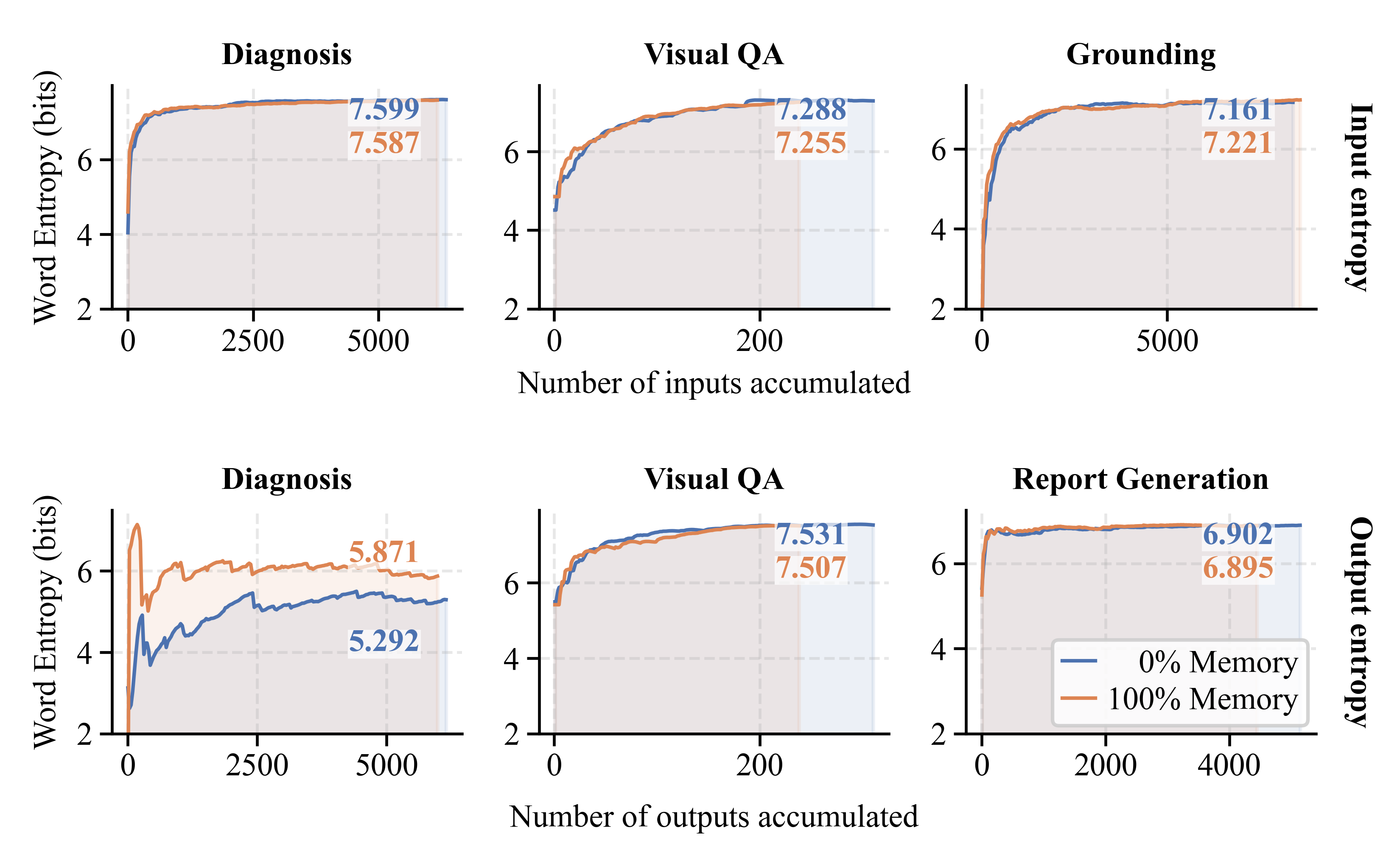}
\caption{\textbf{Corpus-level cumulative entropy: memory yields more diverse diagnostic outputs per call.}
Cumulative word-level Shannon entropy of the concatenated input (top row) and output (bottom row) corpora as tool calls are appended.
The Diagnosis output gap ($+0.58$ bits; $5.29\!\to\!5.87$) shows that memory eliminates near-duplicate diagnostic calls---the retained outputs are markedly less redundant across cases---while all other corpora remain essentially unchanged.}
\label{fig:corpus_entropy_app}
\end{figure*}

\paragraph{Corpus-level cumulative entropy (Figure~\ref{fig:corpus_entropy_app}).}
Per-call entropy cannot detect whether two lexically rich queries are in fact asking the same thing. The corpus-level cumulative entropy---computed by concatenating all call inputs (or outputs) of a tool into a single corpus and tracking its word-level Shannon entropy as calls accumulate---resolves this ambiguity. For the \texttt{Dgn} tool, the memory-augmented agent's output corpus reaches $+0.58$ bits higher than the Baseline ($5.29\!\to\!5.87$), despite issuing fewer calls in total.
Since per-call output entropy is unchanged (Figure~\ref{fig:info_entropy_app}), this pattern suggests that memory mainly changes which calls are issued rather than how individual calls are phrased. The higher corpus-level entropy for \texttt{Dgn} is consistent with fewer near-duplicate diagnostic calls and more differentiated retained outputs across cases.
All other corpora---the \texttt{Dgn} \emph{input} side, together with both inputs and outputs of \texttt{Vqa} and \texttt{Rpt}---remain within $0.07$ bits of the Baseline, confirming that the effect is specific to the diagnostic-reasoning channel where redundancy can be meaningfully removed.

\section{Computational Cost}
\label{sec:overhead}

Table~\ref{tab:overhead} compares the per-case inference-stage computational cost of the memory-free baseline and EATS on ChestAgentBench. EATS reduces the per-case inference-stage time and token usage by $10.1\%$ and $10.5\%$, respectively. This reduction is consistent with the tool-use analysis in Section~\ref{sec:tool_use_analysis}: retrieved experience suppresses low-value tool calls and shortens the reasoning process, which outweighs the additional context introduced by the retrieved references.

\begin{table}[t]
\centering
\small
\caption{Per-case inference-stage computational cost on ChestAgentBench, averaged over $2{,}500$ cases. $\Delta$ is relative to the memory-free baseline.}
\label{tab:overhead}
\setlength{\tabcolsep}{3pt}
\begin{tabular*}{\linewidth}{@{\extracolsep{\fill}}lccc@{}}
\toprule
Per-case metric & w/o Memory & w/ Memory & $\Delta$ \\
\midrule
Time (s)   & 64.1       & 57.6       & $-10.1\%$ \\
Tokens     & $29{,}786$ & $26{,}653$ & $-10.5\%$ \\
\bottomrule
\end{tabular*}
\end{table}

\section{Limitations}
\label{sec:limitations}

Due to the substantial computational and financial cost of repeated tool-mediated agent inference, our empirical validation focuses on widely used chest X-ray benchmarks and a representative set of specialized neural tools. This leaves broader cross-domain validation, such as other medical imaging modalities or clinical specialties, for future work.

We expect the same underlying challenge to arise beyond the CXR setting: specialized neural tools often have case-dependent reliability, and agents still need to decide which tools to invoke for each case. Extending the framework to other domains mainly requires replacing the tool library and datasets, while keeping the memory-augmented tool-selection mechanism unchanged.

\section{Implementation Details}
\label{sec:implementation_details_app}

This section documents the MedRAX system prompt and default tool interface used by the no-memory baseline agent and our memory-augmented variant in the main experiments.

\paragraph{System prompt.}
MedRAX uses a medical-assistant system prompt that instructs the model to rely on its own visual reasoning, use tools as complementary evidence, make multiple tool calls when needed, and critically assess tool outputs. The prompt also informs the model that image paths passed to tools can be simple filenames or partial paths, because the execution system resolves them to the corresponding full paths.

\paragraph{Tool library.}
The main experiments use six-tool list defined in \texttt{benchmark\_medrax\_v11v2.py} line~103 in our \texttt{Code.zip} file (see supplementary material). In the \texttt{API} backend, these tools are exposed through the same-name remote wrappers with substantially the same descriptions.

\begin{description}[leftmargin=1.5em,style=nextline,itemsep=0.45em]
    \item[\texttt{chest\_xray\_report\_generator (Rpt)}.]
    This tool analyzes chest X-ray images and generates comprehensive radiology reports containing detailed findings and impression summaries.
    Input: \texttt{image\_path}, the chest X-ray image path.

    \item[\texttt{chest\_xray\_classifier (Cls)}.]
    This tool analyzes chest X-ray images and classifies them for 18 pathologies, returning predicted probabilities from 0 to 1, where higher values indicate higher likelihood.
    The pathologies are Atelectasis, Cardiomegaly, Consolidation, Edema, Effusion, Emphysema, Enlarged Cardiomediastinum, Fibrosis, Fracture, Hernia, Infiltration, Lung Lesion, Lung Opacity, Mass, Nodule, Pleural Thickening, Pneumonia, and Pneumothorax.
    Input: \texttt{image\_path}, the chest X-ray image path.

    \item[\texttt{chest\_xray\_segmentation (Seg)}.]
    This tool segments chest X-ray images to specified anatomical structures and returns a segmentation visualization with comprehensive metrics.
    Available organs are Left/Right Clavicle, Left/Right Scapula, Left/Right Lung, Left/Right Hilus Pulmonis, Heart, Aorta, Facies Diaphragmatica, Mediastinum, Weasand, and Spine.
    The area estimate is not accurate unless the input has been DICOM.
    Inputs: \texttt{image\_path}; optional \texttt{organs}.

    \item[\texttt{xray\_phrase\_grounding (Grd)}.]
    This tool locates and visualizes specific medical findings in frontal chest X-ray images, returning normalized bounding-box coordinates, a visualization, and confidence metadata.
    Inputs: \texttt{image\_path}; \texttt{phrase}; optional \texttt{max\_new\_tokens}.

    \item[\texttt{chest\_xray\_expert (Dgn)}.]
    This versatile chest X-ray analysis tool supports visual question answering, report generation, abnormality detection, comparative analysis, anatomical description, and clinical interpretation.
    Inputs: \texttt{image\_paths}; \texttt{prompt}; optional \texttt{max\_new\_tokens}.

    \item[\texttt{llava\_med\_qa (Vqa)}.]
    This tool answers questions about biomedical images and general medical questions using LLaVA-Med.
    While it can process chest X-rays, it may not be as reliable for detailed chest X-ray analysis.
    Inputs: \texttt{question}; optional \texttt{image\_path}.
\end{description}

\end{document}